\pdfoutput=1

\documentclass[11pt]{article}

\usepackage[]{naacl2021}

\usepackage{times}
\usepackage{latexsym}
\usepackage{graphicx}
\usepackage{multirow}
\usepackage{multicol}
\usepackage{makecell}
\usepackage{xcolor}
\definecolor{mygreen}{RGB}{34, 139, 34}

\usepackage[T1]{fontenc}

\usepackage[utf8]{inputenc}

\usepackage{microtype}

\title{Explore BiLSTM-CRF-Based Models for Open Relation Extraction}

\author{Tao Ni \\
  Australian National University, Australia \\
  \texttt{u6788943@anu.edu.au} \\\And
  Qing Wang \\
  Australian National University, Australia \\
  \texttt{qing.wang@anu.edu.au} \\\AND
  Gabriela Ferraro \\
  CSIRO Data61, Canberra, Australia \\
  \texttt{gabriela.ferraro@data61.csiro.au} \\}

\begin{document}
\maketitle
\begin{abstract}
Extracting multiple relations from text sentences is still a challenge for current Open Relation Extraction (Open RE) tasks. In this paper, we develop several Open RE models based on the bidirectional LSTM-CRF (BiLSTM-CRF) neural network and different contextualized word embedding methods. We also propose a new tagging scheme to solve overlapping problems and enhance models' performance. From the evaluation results and comparisons between models, we select the best combination of tagging scheme, word embedder, and BiLSTM-CRF network to achieve an Open RE model with a remarkable extracting ability on multiple-relation sentences.
\end{abstract}

\section{Introduction}

\begin{figure}[h]
    \centering
    \includegraphics[width=\linewidth]{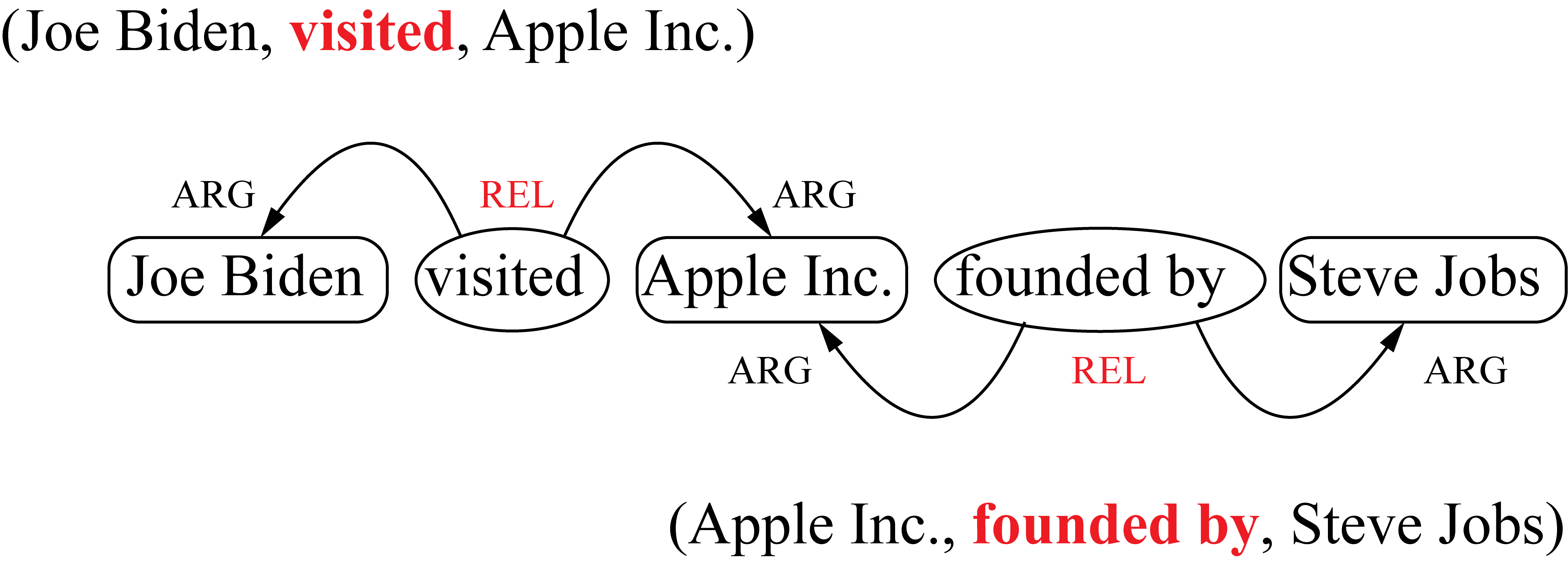}
    \caption{Example of Open Relation Extraction (ARG: Argument, REL: Relation)}
    \label{fig:ore}
\end{figure}

Open Relation Extraction (Open RE) is an important task of Natural Language Processing (NLP), which involves extracting structured relation representations from text sentences. In an Open RE system, relations are expressed by predicates and their arguments. For instance, Figure~\ref{fig:ore} shows that in the example sentence ``Joe Biden visited Apple Inc. founded by Steve Jobs'', ``visited'' is the predicate and ``Joe Biden'' and ``Apple Inc.'' are the arguments of the relation while ``founded by'' indicates another relation between ``Apple Inc.'' and ``Steve Jobs''. Nowadays, Open RE has been applied widely in knowledge-based applications such as question-answering intelligence, dialogue systems, ontology building, and text comprehension \citep{cui2018neural,jia2019hybrid}.

Most of the existing Open RE models such as TextRunner \citep{etzioni2008open}, Reverb \citep{fader2011identifying} and OLLIE \citep{schmitz2012open} use pattern-based matching method. But the hand-crafted patterns highly rely on rule-based or semi-supervised algorithms \citep{jia2019hybrid}. Even though using such hand-crafted patterns can extract relations accurately, it requires much more time and human resources, thereby reducing the efficiency of an Open RE task.

Recurrent Neural Network (RNN) is a popular artificial neural network that exhibits many advantages on sequence-relevant tasks. Many RNN-based models are proposed and the Long Short-term Memory (LSTM) \citep{hochreiter1997long} is widely considered as one of the most reliable and effective RNN networks for building Open RE models.

To better understand the context of sentences, bidirectional LSTM (BiLSTM) has been proposed \citep{huang2015bidirectional}. It handles the input in two directions: one is from the start to the end and another is from the end to the start. This architecture shows excellent performance for prediction. Recently, the BiLSTM-CRF network has been widely applied on sequence tagging tasks such as Named Entity Recognition (NER) and Part-Of-Speech (POS) tagging \citep{jia2019hybrid}. Several state-of-the-art models are used for dynamic word representations such as CoVe \citep{mccann2017learned}, ELMo \citep{Peters:2018} and BERT \citep{devlin2018bert}.

In this study, we build our Open RE models by applying a BiLSTM network as the encoding layer and a Conditional Random Field (CRF) layer for tagging words. We use both pre-trained word embeddings and contextualized word embedders BERT and ELMo to produce word representations and feed them into the BiLSTM-CRF neural network. We also train our BiLSTM-CRF-based models under different tagging 
schemes to find an Open RE model with superior performance.

\section{Related Work}

Generally, there are two main directions for building a supervised Open RE model in the literature. One is to create a sequence tagging model and the other is to design a sequence-to-sequence model.

\noindent \textbf{Sequence Tagging Models.} \citeauthor{stanovsky2018supervised} \citeyearpar{stanovsky2018supervised} introduced the first supervised Open RE model for extracting both predicates and related arguments. They used a BiLSTM network with a Softmax layer to build an RNN model which tags each word in the sentences and extract tuples that contain relations. Further, they also created a large-scale, high-quality labelled corpus to support supervised training.

\citeauthor{jia2019hybrid} \citeyearpar{jia2019hybrid} presented a Hybrid Neural Tagging Model for Open RE task (HNN4ORT). They used a triple embedding layer, the Dual Aware Mechanism and Ordered Neurons LSTM (ON-LSTM) to build the hybrid neural framework \citep{jia2019hybrid}. Moreover, they also proposed the idea about changing the rule of labelling in the dataset to improve model performance.

\noindent \textbf{Sequence-to-sequence Models.} Open RE can also be regarded as a sequence-to-sequence (seq2seq) task based on a machine translation mechanism. \citeauthor{cui2018neural} \citeyearpar{cui2018neural} introduced an encoder-decoder framework with multiple-LSTM layers which can generate sequences with placeholders and extract relation tuples from the output. Consider the example sentence in Figure~\ref{fig:ore}, the output sequence will be ``$\langle arg0 \rangle$Joe Biden$\langle /arg0 \rangle$ $\langle rel \rangle$visited$\langle /rel \rangle$ $\langle arg1 \rangle$Apple Inc.$\langle /arg1 \rangle$ $\langle rel \rangle$founded by$\langle /rel \rangle$ $\langle arg2 \rangle$Steve Jobs$\langle /arg2 \rangle$''. Hence, two relation tuples (Joe Biden, \textbf{visited}, Apple Inc.) and (Apple Inc., \textbf{founded by}, Steve Jobs) can be extracted by identifying relations between placeholders ``$\langle rel \rangle$'' and ``$\langle /rel \rangle$''. Besides, \citeauthor{zhang2017mt} \citeyearpar{zhang2017mt} also built a seq2seq model for joint extractions of relation-argument tuples and extended Open RE tasks to other languages such as Mandarin.

So far, several Open RE models have been proposed which can achieve competitive performance against traditional RE models. However, these Open RE models usually perform well in extracting only one relation per sentence, and they often annotate relations separately. Although they exhibit the great ability to extract single-relation sentences but suffer from poor performance when an input sentence has multiple relations. In addition, few of these models showed the model performance with different word embedding methods especially some advanced contextualized word embedders.

In our study, we use both pre-trained word embeddings and contextualized word embedders BERT \citep{devlin2018bert} and ELMo \citep{Peters:2018} with BiLSTM-CRF network to build Open RE models. We train these models on a high-quality dataset and compare the models' performances to find the best combination of word embedders,  BiLSTM-CRF and parameter settings. In addition, we develop a new tagging scheme to alleviate the drawbacks of the data corpus and it also improves the Open RE models' performance on extracting multiple relations from sentences.

\section{Method}

\subsection{Tagging Scheme}

We define Open RE as a sequence tagging task.
Let $S=(w_1, w_2, ...,w_n)$ be a word sequence (i.e., a sentence) where each $w_i$ is a word, and $T=(t_1, t_2, ...,t_n)$ be a tagging sequence where each $t_i$ is a tag. For tags in $T$, we use \textbf{P-B} and \textbf{P-I} as relation tags to denote predicates, \textbf{A-B} and \textbf{A-I} as argument tags to denote arguments, and \textbf{O} to denote other words. 
For the training corpus, we use a public large-scale labeled dataset - All Words Open IE \footnote{\url{https://github.com/gabrielStanovsky/supervised-oie}} (\verb|AW-OIE|) \citep{stanovsky2018supervised}. They proposed a tagging scheme in where a word sentence may generate several tagging sequences where each sequence has only one relation tag. Each tagging sequence is independent of each other so that they extended the training set to 12952 samples for supervised training.

\begin{figure*}[htbp]
    \centering
    \includegraphics[width=\textwidth]{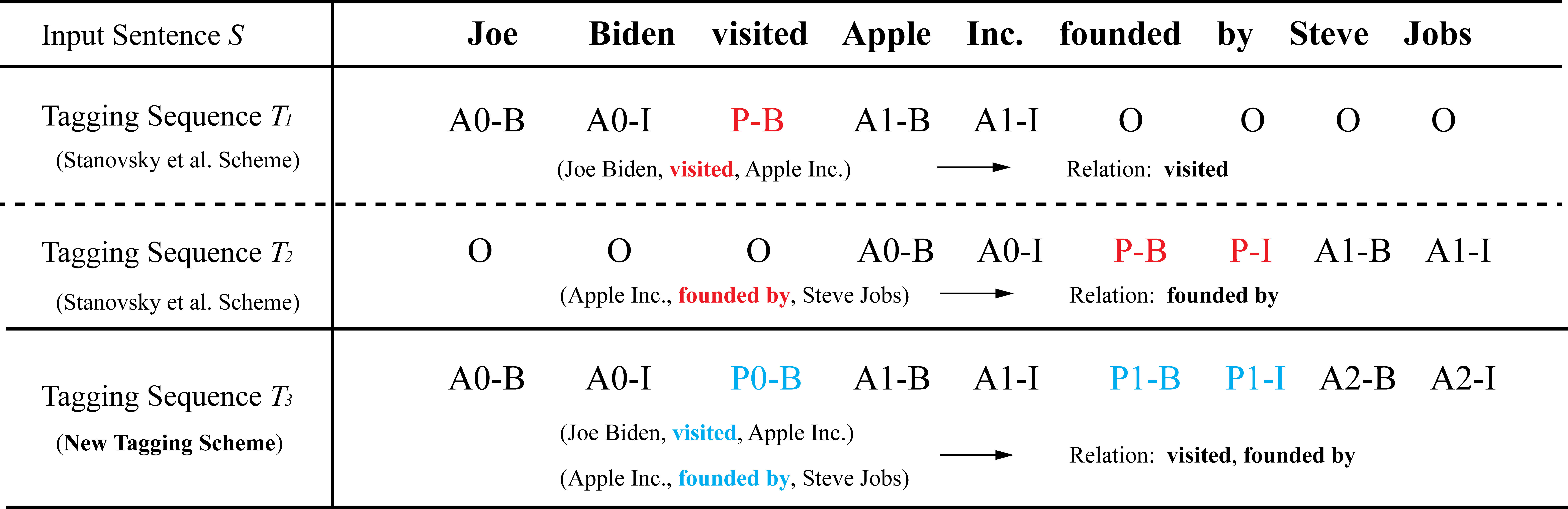}
    \caption{\citeauthor{stanovsky2018supervised} \citeyearpar{stanovsky2018supervised} Tagging Scheme only produces one tag per sentence. For a sentence with multiple relations, there are some overlapping tag representations of one same word across different tagging sequences. In the New Tagging Scheme, relations are annotated orderly so that less \textbf{O} tags are used.}
    \label{fig:taggingscheme}
\end{figure*}

However, this tagging scheme has overlapping problems in multiple-relation sentences. For example, in Figure~\ref{fig:taggingscheme}, relation ``visited'' is denoted as \textbf{P-B} in a tagging sequence $T_1$ but as \textbf{O} in another tagging sequence $T_2$. The overlapping problems cause incomplete and incorrect predictions in the sequence tagging process and impact the model performance. To solve overlapping problems, we propose a new tagging scheme that can automatically produce tagging sequences with multiple relation annotations. In Figure~\ref{fig:taggingscheme}, the tagging sequence $T_3$ is generated by our new tagging scheme and it has multiple relation tags \textbf{P0-B}, \textbf{P1-B} and \textbf{P1-I} based on the order of relations. And we don't need to expand our training data based on the new tagging scheme.

\subsection{Model Architecture}

Our model consists of three layers: an embedding layer (embedder), a bidirectional LSTM (BiLSTM) layer, and a  Conditional Random Field (CRF) layer.

For the embedding layer, we use pre-trained word embeddings like GloVe \citep{pennington2014glove}. We also use two deep contextualized word embedders, BERT and ELMo, to produce the representation of each tag based on its context in a sentence.

Then, the word embedding sequence $X=(x_1, x_2,..., x_n)$ is fed into a BiLSTM layer which is made up by forward and backward LSTM cells. The forward LSTM cells generate the forward hidden states $H_f = (h_{f_1}, h_{f_2},..., h_{f_n})$ as follows \citep{cui2018neural}.

\begin{equation}
    h_{f(t)} = \textbf{LSTM}(x_t, h_{f(t-1)})
\end{equation}

Similarly, the backward LSTM cells generate the backward hidden states $H_b = (h_{b_1}, h_{b_2},..., h_{b_n})$. 

\begin{equation}
    h_{b(t)} = \textbf{LSTM}(x_t, h_{b(t-1)})
\end{equation}

Then the BiLSTM layer combines forward and backward hidden states to produce the complete hidden state sequence $H = (h_1, h_2,..., h_n)$. It also maps hidden state vectors to $k$-dimension vectors where $k$ represents the number of tags in the tagging scheme \citep{dai2019named}. Hence, the BiLSTM layer extracts and combines both the ``past'' and ``future'' features from a sentence.

The CRF layer takes the output of the BiLSTM layer and starts the sequence tagging process. It has the ability to make use of contextualized tag information to produce a better tagging accuracy \citep{huang2015bidirectional}. As a result, we recognize relation tags \textbf{P-B} (New Tagging Scheme: \textbf{P0-B}, \textbf{P1-B}...) from the output tagging sequence to finish the Open RE task.


\section{Experiments}

\subsection{Data}

For the training data, we use the \verb|AW-OIE| data corpus which is derived from \verb|OIE2016| by QA-SRL annotations \citep{he2015question}. We extract a training set with 3300 multiple-relation sentences with 12952 corresponding tagging sequences. Each sentence has about 4 relations. The tags are extended from BIO tagging scheme with a head notation \textbf{A} to denote arguments (\textbf{A0-B}, \textbf{A0-I}) and \textbf{P} to denote relations (\textbf{P-B}, \textbf{P-I}) \citep{stanovsky2018supervised}.

For the testing data, we collect testing sets from previous Open RE works to evaluate the performance of our Open RE model. Table~\ref{tab:testingsets} depicts the three testing sets in the experiments: \verb|AW-OIE|, \verb|Wikipedia| and \verb|NYT| \citep{del2013clausie}. We extract another 634 sentences with 1724 relations from the \verb|AW-OIE| corpus to build the \verb|AW-OIE| testing set. Additionally, the \verb|Wikipedia| testing set contains 376 sentences with 985 relations, and the \verb|NYT| testing set has 258 sentences with 739 relations. The average number of relations within each sentence is close to 3, and these testing sets can be used for evaluating the performance of our Open RE model on multiple-relation extractions.

\begin{table}[h]
\centering
\begin{tabular}{cccc}
\hline
\textbf{Dataset} & \textbf{Sentence} & \textbf{Relation} & \textbf{Avg (R/S)}\\
\hline
AW-OIE & 634 & 1724 & 2.7 \\
Wikipedia & 376 & 985 & 2.6 \\
NYT & 258 & 739 & 2.9 \\
\hline
\end{tabular}
\caption{Profile of testing sets in the experiments. Avg (R/S) means average number of relations per sentence.}
\label{tab:testingsets}
\end{table}

\subsection{Word Embeddings}

In the experiments, we combine the BiLSTM-CRF network with different word embedding methods.

\textbf{GloVe-BiLSTM-CRF}. We build the GloVe-BiLSTM-CRF model by applying the pre-trained word embeddings GloVe \citep{pennington2014glove}. We choose 200-dimension vectors for capturing the word-level semantics.

\textbf{W2V-BiLSTM-CRF}. We also select another 300-dimension Word2Vec \citep{mikolov2013distributed} pre-trained vectors from the Google News dataset to build the W2V-BiLSTM-CRF model.

\textbf{BERT-BiLSTM-CRF}. We use a pre-trained BERT model from Google Research \citep{turc2019} to generate deep contextualized word representations. We choose the BERT-Base, Multilingual Cased version.

\textbf{ELMo-BiLSTM-CRF}. We build the ELMo-BiLSTM-CRF model by using an original pre-trained ELMo model provided by AllenNLP \citep{Peters:2018} as the word embedder.

\subsection{Parameter Settings}

The experimental environment is on the Amazon Web Service (AWS) platform with the setting of 16 vCPU, 2.9 GHz Intel Xeon E5-2666 v3 Processor, RAM. We implement models on PyTorch \citep{paszke2019pytorch} and use RayTune \citep{liaw2018tune} for hyper-parameters tuning.

For both the GloVe-BiLSTM-CRF model and the W2V-BiLSTM-CRF model, we initialize the learning rate as 0.01 with a weight decay of 0.0001. The mini-batch size is set to 50 and we train the models in 100 epochs.

For the BERT-BiLSTM-CRF model and the ELMo-BiLSTM-CRF model, we set most hyper-parameters based on prior work \citep{devlin2018bert, Peters:2018} and only changes token representations of tags. We also train these two models in 100 epochs.

We use Adam \citep{kingma2014adam} optimizer for all models in the experiment. Meanwhile, we train the models under \citeauthor{stanovsky2018supervised} \citeyearpar{stanovsky2018supervised} tagging scheme and the new tagging scheme with the same testing sets to make a fair comparisons.

\subsection{Evaluation}

The evaluation metrics in our experiment are precision, recall and F1-score. The recall values are used to evaluate the rate of how many relations are correctly predicted, Furthermore, we define a new measurement called \textbf{Predicate Matching Score} (PMS) to represent the correct relation extraction rate for evaluating the model performance. The PMS is able to indicate the performance of a model based on different tagging schemes. $Predict(R_i)$ and $True(R_i)$ represent the lists of predicted and true relations for the $i$-th input sentence in the testing set, and $l$ represents the number of all sentences in the testing set.

\begin{equation}
    PMS (\%) = \frac{\sum_{i=1}^{l} |Predict(R_i)|}{\sum_{i=1}^{l} |True(R_i)|}\times 100\%
\end{equation}

Table ~\ref{tab:pms} shows an example of computing PMS. We first remove incorrect predicted relations, stopwords and modal verbs like "could" to guarantee that the predicted relation list is a sublist of true relation list. Then we compare the extracted relations with the true relations and calculate the PMS of the testing set. As sequence tagging tasks always produce non-relation tags, using PMS to assess the Open RE model can mitigate the influence of arguments prediction and focus on the models' performance of relation prediction.

\begin{table*}[h]
\centering
\begin{tabular}{c|c|c}
\hline
    \textbf{Sentences}    & \textbf{Extracted Relations (Predict($R_i$))}  & \textbf{True Relations (True($R_i$))}          \\ \hline
Sent. 1 & {[}"visited"{]}                       & {[}"visited", "founded by"{]}         \\ \hline
Sent. 2 & {[}"participate", "produced"{]}       & {[}"participate", "produced"{]}       \\ \hline
Sent. 3 & {[}"appointed", "could", "proceed"{]} & {[}"appointed", "filed", "proceed"{]} \\ \hline
\textbf{PMS}     & \multicolumn{2}{c}{PMS = $(1 + 2 + 2)/(2 + 2 + 3)\times 100\% = 71.4\%$}                                                    \\ \hline
\end{tabular}
\caption{Example of computing the PMS: Sent.1 incomplete extraction, Sent.2 correctly extraction, Sent.3 incorrect extraction (Remove "could"). So, $|Predict(R_1)|$ = 1, $|Predict(R_2)|$ = 2, $|Predict(R_3)|$ = 2 and $|True(R_1)|$ = 2, $|True(R_2)|$ = 2, $|True(R_3)|$ = 2. We get the PMS = 71.4\%}
\label{tab:pms}
\end{table*}

\section{Results and Discussion}

\subsection{Results}

Table~\ref{tab:result} shows the results of the Open RE models on different testing sets respectively. It also displays the results of NTS-GloVe-BiLSTM-CRF, NTS-W2V-BiLSTM-CRF, NTS-BERT-BiLSTM-CRF and NTS-ELMo-BiLSTM-CRF which represent models that combined with the New Tagging Scheme (NTS). And we also select the results on the same testing sets from previous work proposed by \citeauthor{jia2019hybrid} \citeyearpar{jia2019hybrid} as the baseline.

\begin{table*}
\centering
\begin{tabular}{l|c|c|c|c|c|c|c|c|c|c|c|c}
\hline
                     & \multicolumn{4}{c|}{\textbf{AW-OIE}} & \multicolumn{4}{c|}{\textbf{Wikipedia}} & \multicolumn{4}{c}{\textbf{NYT}} \\ \hline
\textbf{Model}       & P       & R       & F1      & PMS\%    & P        & R        & F1      & PMS\%     & P      & R      & F1     & PMS\%    \\ \hline
Reverb    & 0.64    & 0.16    & 0.26    & /    & 0.77     & 0.21     & 0.33    & /     & 0.56   & 0.14   & 0.23   & /    \\
OLLIE    & 0.98    & 0.24    & 0.38    & /    & 0.99     & 0.28     & 0.44    & /     & 0.98   & 0.25   & 0.40   & /    \\
OpenIE4    & 0.79    & 0.33    & 0.47    & /    & 0.77     & 0.34     & 0.47    & /     & 0.80   & 0.34   & 0.48   & /    \\
HNN4ORT    & 0.85    & 0.75    & 0.80    & /    & 0.84     & 0.76     & 0.79    & /     & 0.83   & 0.68   & 0.75   & /    \\ \hline \hline
GloVe    & 0.88    & 0.44    & 0.57    & 46.1    & 0.89     & 0.44     & 0.59    & 52.8     & 0.86   & 0.39   & 0.53   & 58.9    \\ 
W2V      & 0.86    & 0.21    & 0.29    & 28.9    & 0.87     & 0.22     & 0.31    & 32.2     & 0.85   & 0.23   & 0.32   & 35.3    \\ 
BERT       & 0.88    & 0.74    & 0.80    & 79.2    & 0.89     & 0.75     & 0.81   & 81.1     & 0.86   & 0.77   & 0.82   & 81.9    \\ 
ELMo      & 0.85    & 0.49    & 0.61    & 55.1    & 0.86     & 0.51     & 0.63    & 57.2     & 0.88   & 0.53   & 0.65   & 59.5    \\ \hline
NTS-GloVe & 0.89    & \textbf{0.87}    & \textbf{0.88}    & \textbf{93.6}   & 0.91     & \textbf{0.88}     & \textbf{0.89}    & \textbf{94.8}    & 0.86   & \textbf{0.83}   & \textbf{0.84}   & \textbf{94.1}   \\ 
NTS-W2V   & 0.55    & 0.48    & 0.49    & 84.3    & 0.55     & 0.47     & 0.48    & 85.5     & 0.54   & 0.50   & 0.51   & 83.4    \\
NTS-BERT   & 0.92    & \textbf{0.89}    & \textbf{0.90}    & \textbf{94.3}    & 0.93     & \textbf{0.90}     & \textbf{0.91}    & \textbf{95.5}     & 0.90   & \textbf{0.86}   & \textbf{0.87}   & \textbf{94.3}    \\
NTS-ELMo   & 0.89    & 0.69    & 0.77    & 83.6    & 0.90     & 0.72     & 0.79    & 85.5     & 0.88   & 0.70   & 0.78    & 84.3    \\ \hline
\end{tabular}%
\caption{Evaluation of existing Open RE models (Reverb, OLLIE, OpenIE4, HNN4ORT) \citep{jia2019hybrid} and our Embedder-BiLSTM-CRF models on AW-OIE, Wikipedia and NYT testing sets. Bold values indicate a better performance compared with other models.}
\label{tab:result}
\end{table*}


\begin{table*}[h]
\centering
\begin{tabular}{l|l}
\hline
\multicolumn{1}{c|}{\textbf{Sent. 1}} & While \textbf{pursuing} his MFA at Columbia University, Scieszka \textbf{painted} apartments                                                      \\ \hline
\multirow{2}{*}{GloVe}        & While [pursuing]\textcolor{red}{$_{P-B}$} his MFA at Columbia University, Scieszka [painted]\textcolor{red}{$_{P-B}$} apartments                                  \\ 
                              & While [pursuing]\textcolor{cyan}{$_{P1-B}$} his MFA at Columbia University, Scieszka [painted]\textcolor{cyan}{$_{P0-B}$} apartments                                \\ \hline
\multirow{2}{*}{W2V}          & While [pursuing]\textcolor{mygreen}{$_{O}$} his MFA at Columbia University, Scieszka [painted]\textcolor{red}{$_{P-B}$} apartments                                    \\
                              & While [pursuing]\textcolor{mygreen}{$_{O}$} his MFA at Columbia University, Scieszka [painted]\textcolor{cyan}{$_{P0-B}$} apartments                                \\ \hline
\multirow{2}{*}{BERT}         & While [pursuing]\textcolor{red}{$_{P-B}$} his MFA at Columbia University, Scieszka [painted]\textcolor{red}{$_{P-B}$} apartments                                  \\
                              & While [pursuing]\textcolor{cyan}{$_{P1-B}$} his MFA at Columbia University, Scieszka [painted]\textcolor{cyan}{$_{P0-B}$} apartments                                \\ \hline
\multirow{2}{*}{ELMo}         & While [pursuing]\textcolor{mygreen}{$_{O}$} his MFA at Columbia University, Scieszka [painted]\textcolor{red}{$_{P-B}$} apartments                                    \\ 
                              & While [pursuing]\textcolor{cyan}{$_{P1-B}$} his MFA at Columbia University, Scieszka [painted]\textcolor{cyan}{$_{P0-B}$} apartments                                \\ \hline \hline
\textbf{Sent. 2}                       & \makecell[l]{Mrs. Marcos has not \textbf{admitted} that she \textbf{filed} any documents such as those \\ \textbf{sought by} the government}                                             \\ \hline
\multirow{2}{*}{GloVe}        & \makecell[l]{Mrs. Marcos has not [admitted]\textcolor{red}{$_{P-B}$} that she [filed]\textcolor{mygreen}{$_{O}$} any documents such as those \\ {[sought]\textcolor{red}{$_{P-B}$}} [by]\textcolor{mygreen}{$_{O}$} the government}               \\
                              & \makecell[l]{Mrs. Marcos has not [admitted]\textcolor{cyan}{$_{P0-B}$} that she [filed]\textcolor{mygreen}{$_{O}$} any documents such as those \\ {[sought]\textcolor{cyan}{$_{P1-B}$}} [by]\textcolor{mygreen}{$_{O}$} the government}               \\ \hline
\multirow{2}{*}{W2V}          & \makecell[l]{Mrs. Marcos has not [admitted]\textcolor{red}{$_{P-B}$} that she [filed]\textcolor{mygreen}{$_{O}$} any documents such as those \\ {[sought]\textcolor{mygreen}{$_{O}$} [by]\textcolor{mygreen}{$_{O}$}} the government}                   \\
                              & \makecell[l]{Mrs. Marcos has not [admitted]\textcolor{cyan}{$_{P0-B}$} that she [filed]\textcolor{mygreen}{$_{O}$} any documents such as those \\ {[sought]\textcolor{cyan}{$_{P1-B}$}} [by]\textcolor{mygreen}{$_{O}$} the government}               \\ \hline
\multirow{2}{*}{BERT}         & \makecell[l]{Mrs. Marcos has not [admitted]\textcolor{red}{$_{P-B}$} that she [filed]\textcolor{red}{$_{P-B}$} any documents such as those \\ {[sought]\textcolor{red}{$_{P-B}$}} [by]\textcolor{mygreen}{$_{O}$} the government}               \\
                              & \makecell[l]{Mrs. Marcos has not [admitted]\textcolor{cyan}{$_{P0-B}$} that she [filed]\textcolor{cyan}{$_{P1-B}$} any documents such as those \\ {[sought]\textcolor{cyan}{$_{P2-B}$}} [by]\textcolor{cyan}{$_{P2-I}$} the government} \\ \hline
\multirow{2}{*}{ELMo}         & \makecell[l]{Mrs. Marcos has not [admitted]\textcolor{red}{$_{P-B}$} that she [filed]\textcolor{mygreen}{$_{O}$} any documents such as those \\ {[sought]\textcolor{red}{$_{P-B}$}} [by]\textcolor{mygreen}{$_{O}$} the government}                 \\
                              & \makecell[l]{Mrs. Marcos has not [admitted]\textcolor{cyan}{$_{P0-B}$} that she [filed]\textcolor{cyan}{$_{P1-B}$} any documents such as those \\ {[sought]\textcolor{cyan}{$_{P2-B}$}} [by]\textcolor{mygreen}{$_{O}$} the government}         \\ \hline
\end{tabular}
\caption{Examples of experimental output of our embedder-BiLSTM-CRF models using two tagging schemes. Bold words in the sentences are relations. Red tags are relation tags of \citeauthor{stanovsky2018supervised} \citeyearpar{stanovsky2018supervised} tagging scheme and light blue tags are relation tags of the new tagging scheme. Green tags reflect mispredicted tags.}
\label{tab:extractresults}
\end{table*}

\subsection{Discussion}

We investigate the experimental results and observe that GloVe-BiLSTM-CRF, W2V-BiLSTM-CRF and ELMo-BiLSTM-CRF models all have high precision with relatively low recall and PMS in the tagging scheme of \citeauthor{stanovsky2018supervised} \citeyearpar{stanovsky2018supervised}. The BERT-BiLSTM-CRF model performs well but the results are not competitive enough. Table~\ref{tab:extractresults} provides several RE examples from the experiments. The high precision values result from a small number of false-positive (FP) predictions so that precision is unable to precisely reflect the model performance. The low recall and PMS values indicate the extraction ability is poor because of the existence of incomplete and incorrect relation extractions. We discover that too many \textbf{O} tags in \citeauthor{stanovsky2018supervised} \citeyearpar{stanovsky2018supervised} tagging scheme causes overlapping problems in tagging sequence, leading to the poor recall and PMS values of these models.

Table~\ref{tab:result} also shows the evaluation results of these models based on the new tagging scheme. We find both recall and PMS values have increased considerably in these models. NTS-GloVe-BiLSTM-CRF model and NTS-BERT-BiLSTM-CRF model have remarkable performance i.e., nearly 94\% relations are correctly extracted from the testing sets. BERT-BiLSTM-CRF has better results than GloVe-BiLSTM-CRF in the tagging scheme of \citeauthor{stanovsky2018supervised} \citeyearpar{stanovsky2018supervised}. Nevertheless, from the second example in Table~\ref{tab:extractresults}, NTS-BERT-BiLSTM-CRF can the extract relation phrase "sought by" completely but NTS-GloVe-BiLSTM-CRF can only identify the word "sought". We believe the size of testing sets is one limitation that makes the results from NTS-GloVe-BiLSTM-CRF and NTS-BERT-BiLSTM-CRF very close.

We also notice the increase in recall and PMS from NTS-W2V-BiLSTM-CRF and NTS-ELMo-BiLSTM-CRF. Both models perform better than previous works like Reverb, OLLIE and OpenIE4. However, their performances are not competitive when compared with other advanced Open RE models such as HNN4ORT \citep{jia2019hybrid}. Meanwhile, for NTS-W2V-BiLSTM-CRF, there is a decrease of precision from approximately 0.88 to 0.55 which reflects the existence of a large number of FP predictions under the new tagging scheme.

According to the evaluation results, we find that the NTS-BERT-BiLSTM-CRF outperforms other BiLSTM-CRF based models. Our new tagging scheme can enhance the Open RE models' performance. The NTS-BERT-BiLSTM-CRF displays excellent performance on the Open RE task and especially achieves over 94\% correctly extractions in multiple-relation sentences.

\section{Conclusion}

In this paper, we propose and explore several BiLSTM-CRF based models for open relation extraction. We also introduce a new tagging scheme to solve overlapping problems and improve the performance of models. Experimental results indicate that NTS-BERT-BiLSTM-CRF and NTS-GloVe-BiLSTM-CRF perform well in correctly and completely extracting multiple relations from sentences.

In future work, we are going to explore joint extraction methods~\cite{zheng2017joint, ni2021simple, ni2023eavesdropping, ni2023recovering, ni2023exploiting, ni2023xporter, ni2023uncovering, chen2022swipepass, ni2024sensor, song2023emma} to extract both relations and entities. Furthermore, we will expand datasets and models to work on some cross-lingual extraction \citep{zhang2017mt} tasks such as extracting relations from other languages such as French, Latin and Mandarin.

\bibliography{anthology,custom}

\begin{thebibliography}{31}
\expandafter\ifx\csname natexlab\endcsname\relax\def\natexlab#1{#1}\fi

\bibitem[{Chen et~al.(2022)Chen, Ni, Xu, and Gu}]{chen2022swipepass}
Yongliang Chen, Tao Ni, Weitao Xu, and Tao Gu. 2022.
\newblock Swipepass: Acoustic-based second-factor user authentication for
  smartphones.
\newblock \emph{Proceedings of the ACM on Interactive, Mobile, Wearable and
  Ubiquitous Technologies}, 6(3):1--25.

\bibitem[{Cui et~al.(2018)Cui, Wei, and Zhou}]{cui2018neural}
Lei Cui, Furu Wei, and Ming Zhou. 2018.
\newblock Neural open information extraction.
\newblock \emph{arXiv preprint arXiv:1805.04270}.

\bibitem[{Dai et~al.(2019)Dai, Wang, Ni, Li, Li, and Bai}]{dai2019named}
Zhenjin Dai, Xutao Wang, Pin Ni, Yuming Li, Gangmin Li, and Xuming Bai. 2019.
\newblock Named entity recognition using bert bilstm crf for chinese electronic
  health records.
\newblock In \emph{2019 12th International Congress on Image and Signal
  Processing, BioMedical Engineering and Informatics (CISP-BMEI)}, pages 1--5.
  IEEE.

\bibitem[{Del~Corro and Gemulla(2013)}]{del2013clausie}
Luciano Del~Corro and Rainer Gemulla. 2013.
\newblock Clausie: clause-based open information extraction.
\newblock In \emph{Proceedings of the 22nd international conference on World
  Wide Web}, pages 355--366.

\bibitem[{Devlin et~al.(2018)Devlin, Chang, Lee, and
  Toutanova}]{devlin2018bert}
Jacob Devlin, Ming-Wei Chang, Kenton Lee, and Kristina Toutanova. 2018.
\newblock Bert: Pre-training of deep bidirectional transformers for language
  understanding.
\newblock \emph{arXiv preprint arXiv:1810.04805}.

\bibitem[{Etzioni et~al.(2008)Etzioni, Banko, Soderland, and
  Weld}]{etzioni2008open}
Oren Etzioni, Michele Banko, Stephen Soderland, and Daniel~S Weld. 2008.
\newblock Open information extraction from the web.
\newblock \emph{Communications of the ACM}, 51(12):68--74.

\bibitem[{Fader et~al.(2011)Fader, Soderland, and
  Etzioni}]{fader2011identifying}
Anthony Fader, Stephen Soderland, and Oren Etzioni. 2011.
\newblock Identifying relations for open information extraction.
\newblock In \emph{Proceedings of the 2011 conference on empirical methods in
  natural language processing}, pages 1535--1545.

\bibitem[{He et~al.(2015)He, Lewis, and Zettlemoyer}]{he2015question}
Luheng He, Mike Lewis, and Luke Zettlemoyer. 2015.
\newblock Question-answer driven semantic role labeling: Using natural language
  to annotate natural language.
\newblock In \emph{Proceedings of the 2015 conference on empirical methods in
  natural language processing}, pages 643--653.

\bibitem[{Hochreiter and Schmidhuber(1997)}]{hochreiter1997long}
Sepp Hochreiter and J{\"u}rgen Schmidhuber. 1997.
\newblock Long short-term memory.
\newblock \emph{Neural computation}, 9(8):1735--1780.

\bibitem[{Huang et~al.(2015)Huang, Xu, and Yu}]{huang2015bidirectional}
Zhiheng Huang, Wei Xu, and Kai Yu. 2015.
\newblock Bidirectional lstm-crf models for sequence tagging.
\newblock \emph{arXiv preprint arXiv:1508.01991}.

\bibitem[{Jia and Xiang(2019)}]{jia2019hybrid}
Shengbin Jia and Yang Xiang. 2019.
\newblock Hybrid neural tagging model for open relation extraction.
\newblock \emph{arXiv}, pages arXiv--1908.

\bibitem[{Kingma and Ba(2014)}]{kingma2014adam}
Diederik~P Kingma and Jimmy Ba. 2014.
\newblock Adam: A method for stochastic optimization.
\newblock \emph{arXiv preprint arXiv:1412.6980}.

\bibitem[{Liaw et~al.(2018)Liaw, Liang, Nishihara, Moritz, Gonzalez, and
  Stoica}]{liaw2018tune}
Richard Liaw, Eric Liang, Robert Nishihara, Philipp Moritz, Joseph~E Gonzalez,
  and Ion Stoica. 2018.
\newblock Tune: A research platform for distributed model selection and
  training.
\newblock \emph{arXiv preprint arXiv:1807.05118}.

\bibitem[{McCann et~al.(2017)McCann, Bradbury, Xiong, and
  Socher}]{mccann2017learned}
Bryan McCann, James Bradbury, Caiming Xiong, and Richard Socher. 2017.
\newblock Learned in translation: Contextualized word vectors.
\newblock In \emph{Advances in Neural Information Processing Systems}, pages
  6294--6305.

\bibitem[{Mikolov et~al.(2013)Mikolov, Sutskever, Chen, Corrado, and
  Dean}]{mikolov2013distributed}
Tomas Mikolov, Ilya Sutskever, Kai Chen, Greg~S Corrado, and Jeff Dean. 2013.
\newblock Distributed representations of words and phrases and their
  compositionality.
\newblock \emph{Advances in neural information processing systems},
  26:3111--3119.

\bibitem[{Ni(2024)}]{ni2024sensor}
Tao Ni. 2024.
\newblock Sensor security in virtual reality: Exploration and mitigation.
\newblock In \emph{Proceedings of the 22nd Annual International Conference on
  Mobile Systems, Applications and Services}, pages 758--759.

\bibitem[{Ni et~al.(2021)Ni, Chen, Song, and Xu}]{ni2021simple}
Tao Ni, Yongliang Chen, Keqi Song, and Weitao Xu. 2021.
\newblock A simple and fast human activity recognition system using radio
  frequency energy harvesting.
\newblock In \emph{Adjunct Proceedings of the 2021 ACM International Joint
  Conference on Pervasive and Ubiquitous Computing and Proceedings of the 2021
  ACM International Symposium on Wearable Computers}, pages 666--671.

\bibitem[{Ni et~al.(2023{\natexlab{a}})Ni, Chen, Xu, Xue, and
  Zhao}]{ni2023xporter}
Tao Ni, Yongliang Chen, Weitao Xu, Lei Xue, and Qingchuan Zhao.
  2023{\natexlab{a}}.
\newblock Xporter: A study of the multi-port charger security on privacy
  leakage and voice injection.
\newblock In \emph{Proceedings of the 29th Annual International Conference on
  Mobile Computing and Networking}, pages 1--15.

\bibitem[{Ni et~al.(2023{\natexlab{b}})Ni, Lan, Wang, Zhao, and
  Xu}]{ni2023eavesdropping}
Tao Ni, Guohao Lan, Jia Wang, Qingchuan Zhao, and Weitao Xu.
  2023{\natexlab{b}}.
\newblock Eavesdropping mobile app activity via $\{$Radio-Frequency$\}$ energy
  harvesting.
\newblock In \emph{Proceedings of the 32nd USENIX Security Symposium (USENIX
  Security 23)}, pages 3511--3528.

\bibitem[{Ni et~al.(2023{\natexlab{c}})Ni, Li, Zhang, Zuo, Wang, Xu, Luo, and
  Zhao}]{ni2023exploiting}
Tao Ni, Jianfeng Li, Xiaokuan Zhang, Chaoshun Zuo, Wubing Wang, Weitao Xu,
  Xiapu Luo, and Qingchuan Zhao. 2023{\natexlab{c}}.
\newblock Exploiting contactless side channels in wireless charging power banks
  for user privacy inference via few-shot learning.
\newblock In \emph{Proceedings of the 29th Annual International Conference on
  Mobile Computing and Networking}, pages 1--15.

\bibitem[{Ni et~al.(2023{\natexlab{d}})Ni, Zhang, and Zhao}]{ni2023recovering}
Tao Ni, Xiaokuan Zhang, and Qingchuan Zhao. 2023{\natexlab{d}}.
\newblock Recovering fingerprints from in-display fingerprint sensors via
  electromagnetic side channel.
\newblock In \emph{Proceedings of the 2023 ACM SIGSAC Conference on Computer
  and Communications Security}, pages 253--267.

\bibitem[{Ni et~al.(2023{\natexlab{e}})Ni, Zhang, Zuo, Li, Yan, Wang, Xu, Luo,
  and Zhao}]{ni2023uncovering}
Tao Ni, Xiaokuan Zhang, Chaoshun Zuo, Jianfeng Li, Zhenyu Yan, Wubing Wang,
  Weitao Xu, Xiapu Luo, and Qingchuan Zhao. 2023{\natexlab{e}}.
\newblock Uncovering user interactions on smartphones via contactless wireless
  charging side channels.
\newblock In \emph{Proceedings of the IEEE Symposium on Security and Privacy
  (SP)}, pages 3399--3415. IEEE.

\bibitem[{Paszke et~al.(2019)Paszke, Gross, Massa, Lerer, Bradbury, Chanan,
  Killeen, Lin, Gimelshein, Antiga et~al.}]{paszke2019pytorch}
Adam Paszke, Sam Gross, Francisco Massa, Adam Lerer, James Bradbury, Gregory
  Chanan, Trevor Killeen, Zeming Lin, Natalia Gimelshein, Luca Antiga, et~al.
  2019.
\newblock Pytorch: An imperative style, high-performance deep learning library.
\newblock In \emph{Advances in neural information processing systems}, pages
  8026--8037.

\bibitem[{Pennington et~al.(2014)Pennington, Socher, and
  Manning}]{pennington2014glove}
Jeffrey Pennington, Richard Socher, and Christopher~D Manning. 2014.
\newblock Glove: Global vectors for word representation.
\newblock In \emph{Proceedings of the 2014 conference on empirical methods in
  natural language processing (EMNLP)}, pages 1532--1543.

\bibitem[{Peters et~al.(2018)Peters, Neumann, Iyyer, Gardner, Clark, Lee, and
  Zettlemoyer}]{Peters:2018}
Matthew~E. Peters, Mark Neumann, Mohit Iyyer, Matt Gardner, Christopher Clark,
  Kenton Lee, and Luke Zettlemoyer. 2018.
\newblock Deep contextualized word representations.
\newblock In \emph{Proc. of NAACL}.

\bibitem[{Schmitz et~al.(2012)Schmitz, Soderland, Bart, Etzioni
  et~al.}]{schmitz2012open}
Michael Schmitz, Stephen Soderland, Robert Bart, Oren Etzioni, et~al. 2012.
\newblock Open language learning for information extraction.
\newblock In \emph{Proceedings of the 2012 Joint Conference on Empirical
  Methods in Natural Language Processing and Computational Natural Language
  Learning}, pages 523--534.

\bibitem[{Song et~al.(2023)Song, Ni, Song, and Xu}]{song2023emma}
Keqi Song, Tao Ni, Linqi Song, and Weitao Xu. 2023.
\newblock Emma: An accurate, efficient, and multi-modality strategy for
  autonomous vehicle angle prediction.
\newblock \emph{Intelligent and Converged Networks}, 4(1):41--49.

\bibitem[{Stanovsky et~al.(2018)Stanovsky, Michael, Zettlemoyer, and
  Dagan}]{stanovsky2018supervised}
Gabriel Stanovsky, Julian Michael, Luke Zettlemoyer, and Ido Dagan. 2018.
\newblock Supervised open information extraction.
\newblock In \emph{Proceedings of the 2018 Conference of the North American
  Chapter of the Association for Computational Linguistics: Human Language
  Technologies, Volume 1 (Long Papers)}, pages 885--895.

\bibitem[{Turc et~al.(2019)Turc, Chang, Lee, and Toutanova}]{turc2019}
Iulia Turc, Ming-Wei Chang, Kenton Lee, and Kristina Toutanova. 2019.
\newblock Well-read students learn better: On the importance of pre-training
  compact models.
\newblock \emph{arXiv preprint arXiv:1908.08962v2}.

\bibitem[{Zhang et~al.(2017)Zhang, Duh, and Van~Durme}]{zhang2017mt}
Sheng Zhang, Kevin Duh, and Benjamin Van~Durme. 2017.
\newblock Mt/ie: Cross-lingual open information extraction with neural
  sequence-to-sequence models.
\newblock In \emph{Proceedings of the 15th Conference of the European Chapter
  of the Association for Computational Linguistics: Volume 2, Short Papers},
  pages 64--70.

\bibitem[{Zheng et~al.(2017)Zheng, Wang, Bao, Hao, Zhou, and
  Xu}]{zheng2017joint}
Suncong Zheng, Feng Wang, Hongyun Bao, Yuexing Hao, Peng Zhou, and Bo~Xu. 2017.
\newblock Joint extraction of entities and relations based on a novel tagging
  scheme.
\newblock \emph{arXiv preprint arXiv:1706.05075}.

\end{thebibliography}
\bibliographystyle{acl_natbib}

\end{document}